\documentclass{article}

\usepackage[final]{nips_2016}

\usepackage[utf8x]{inputenc} 
\usepackage{amsfonts,amsmath,amsthm,amssymb}       
\usepackage{nicefrac}       
\usepackage{dsfont}
\usepackage{microtype}      
\usepackage{booktabs}       
\usepackage{bbm,graphicx}
\usepackage{tikz}
\RequirePackage{hypernat}
\RequirePackage[colorlinks=true,citecolor=red,linkcolor=blue,urlcolor=blue]{hyperref}
\usepackage{url}            

\graphicspath{{figures/}}

\newcommand{\E}{\mathbb{E}}
\newcommand{\R}{\mathbb{R}}

\newcommand{\A}{\mathcal{A}}
\newcommand{\D}{\mathcal{D}}

\newcommand{\norm}[1]{\left\Vert #1\right\Vert}

\newcommand{\ind}{\mathds{1}}
\newcommand{\paren}[1]{\left( #1 \right)}
\newcommand{\croch}[1]{\left[\, #1 \,\right]}
\newcommand{\acc}[1]{\left\{ #1 \right\}}
\newcommand{\abs}[1]{\left| #1 \right|}
\newcommand{\scal}[1]{\left\langle #1 \right\rangle_2}
\newcommand{\eucl}[1]{\abs{ #1 }_2}
\newcommand{\N}{\mathbb{N}}
\renewcommand{\P}{\mathbb{P}}
\newcommand{\Rh}{\widehat{R}}
\newcommand{\Y}{\mathcal{Y}}

\newtheorem{postita}{Post-it}

\renewcommand{\L}{\mathcal{L}}
\renewcommand{\S}{\mathcal{S}}

\newtheorem{thm}{Theorem}
\newtheorem{defn}{Definition}
\newtheorem{lem}{Lemma}
\newtheorem{prop}{Proposition}
\newtheorem{cor}{Corollary}

\allowdisplaybreaks

\title{
Stability revisited: new generalisation bounds for the Leave-one-Out
}

\author{
	Alain Celisse\thanks{\url{http://math.univ-lille1.fr/\textasciitilde celisse/}} \\
	Université de Lille \& Inria\\
	Lille, France\\
	\texttt{alain.celisse@math.univ-lille1.fr} \\
	\And
	Benjamin Guedj\thanks{\url{https://bguedj.github.io}} \\
	Inria\\
	Lille, France\\
	\texttt{benjamin.guedj@inria.fr} \\
}

\begin{document}
	
	\maketitle
	\begin{abstract}
		The present paper provides a new generic strategy leading to non-asymptotic theoretical guarantees on the Leave-one-Out procedure applied to a broad class of learning algorithms.
		This strategy relies on two main ingredients: the new notion of $L^q$ stability, and the strong use of moment inequalities. $L^q$ stability extends the ongoing notion of hypothesis stability while remaining weaker than the uniform stability.
		It leads to new PAC exponential generalisation bounds for Leave-one-Out under mild assumptions. In the literature, such bounds are available only for uniform stable algorithms under boundedness for instance.
		Our generic strategy is applied to the Ridge regression algorithm as a first step.
	\end{abstract}
	
	\section{Introduction}

	A massive variety of learning algorithms rely upon unknown parameters that crucially influence the final statistical performance (such as Lasso, Ridge, \dots).
	Cross-validation (CV) procedures are among the most popular data-driven approaches used to assess the performance of estimators, and calibrate their unknown parameters. We refer for instance to \cite{ArlotCelisse_2010} for a survey on CV procedures. Among them, the Leave-one-Out \citep[LoO,][]{Sto:1974} procedure is fairly intuitive, hence widely used.
	Yet its popularity contrasts with the few theoretical results often available in specific settings and derived at the price of strong assumptions \citep[such as boundedness in][Example 3]{BE2002}.
	
	The present paper has the ambition to provide non-asymptotic theoretical guarantees on the LoO procedure.
	We propose a generic strategy to consistently analyse the LoO estimator for learning algorithms. This strategy is based on two ingredients: stability and moment inequalities, which provide concentration results when combined \citep[see for example][for an extensive review]{boucheron2013concentration}. Such concentration results are then precious to derive generalisation bounds, \emph{i.e.}, upper bounds on the discrepancy between the LoO estimator and its prediction error (see for instance \citealp{mcdiarmid1989method} and \citealp{devroye1991exponential} for papers in that direction).

	The notion of stability has first been introduced by \cite{DeWa79} and further studied for instance by \cite{KearnsRon_1999} and \cite{BE2002}.
	This concept has emerged as an effective measure of the "smoothness" of a learning algorithm with respect to its input data.
	For an introduction to stability and connections with other topics such as reproducibility, see \cite{yu2013stability}.
	Over the past decades, the use of stability to derive generalisation bounds has received much attention in the statistical and machine learning community.
	Existing results rely upon stability assumptions such as the \emph{hypothesis} or \emph{uniform stability}.
	For instance, hypothesis stability is used by \citet[][Eq.~(7)]{DeWa79} to derive an upper bound of order 2 moments of LoO for the $k$-nearest neighbors algorithm.
	%
	%
	The stronger uniform stability \citep[][Definition~6]{BE2002} enables to provide a PAC exponential bound for the LoO estimator.

	Further insightful analyses of various notions of stability can be found in
	\cite{KN2002}, \cite{EvgeniouPontilElisseeff_2004}, \cite{ElisseeffEvgeniouPontil_2005},
	\cite{rakhlin2005stability}, \cite{mukherjee2006learning}, \cite{shalev2010learnability}, \cite{kale2011cross}, \cite{kumar2013near} and \cite{villa2013learnability} to name but a few.
	
	\paragraph{Our main contributions.} 
	The present paper introduces a generic strategy to derive new generalisation bounds for the LoO estimator applied to a broad family of learning algorithms.
	This strategy relies on: $(i)$ a new stability assumption that generalises the existing hypothesis stability \citep[][Definition 3]{BE2002} while remaining weaker than uniform stability \citep[][Definition 6]{BE2002}, and $(ii)$ moment inequalities. Combining those two ingredients leads to PAC generalisation bounds.
	For the sake of brevity, we illustrate this strategy by focusing on the Ridge regression algorithm. As part of our contributions, we develop a thorough analysis of the LoO estimator applied to the Ridge regression and obtain generalisation bounds under $L^q$ stability (\autoref{thm.exponential.inequality.LoO.prediction.error1} and \autoref{thm.exponential.inequality.LoO.prediction.error2}) matching state-of-the-art results earlier established by \citet[][Example 3]{BE2002} under the stronger notion of uniform stability. Let us stress though that the proposed strategy is in no way limited to this algorithm and calls for future work to extend it to other algorithms.

    The paper is organised as follows: \autoref{sec:stab} contains our notion of $L^q$ stability for learning algorithms. In particular, we provide an upper bound on the $L^q$ stability of the Ridge regression algorithm (\autoref{thm.Ridge.regression.stability}). \autoref{sec:moments} establishes generalisation bounds in terms of moment inequalities for LoO (\autoref{thm.moments.LoO.prediction.error}). This allows for PAC exponential generalisation bounds in  \autoref{sec:expo}, which is the main achievement of the paper. Specific results in \autoref{thm.exponential.inequality.LoO.prediction.error1} and \autoref{thm.exponential.inequality.LoO.prediction.error2} for the Ridge regression algorithm are also provided. The paper closes with some perspectives in \autoref{sec:discu}, and \autoref{sec:appendix} wraps up technical results.

	\section{Stability of learning algorithms}\label{sec:stab}
	
	The main purpose of the present section is to introduce a generalisation of the notion of $L^1$ stability, also called \emph{hypothesis stability} \citep[][]{DeWa79}, to the higher order $L^q$ stability with $q\geq 2$.
	In particular this new notion turns out to be useful to derive PAC generalisation bounds for the LoO estimator of various learning algorithms (see \autoref{sec:moments} and \autoref{sec:expo}).

	\subsection{Framework and notation}
	
	In what follows, $\A$ denotes a learning algorithm (see \autoref{sec.examples.stable.learning.algorithms} for examples).
	From a training sample $\D = \paren{ Z_1,\ldots,Z_n}\in(\mathcal{X}\times\Y)^{n}$ of $n$ \emph{independent and identically distributed} random variables with $Z_i=(X_i,Y_i)\sim P$ (unknown), $\A$ outputs an estimator $\A(\D):\ \mathcal{X} \subset \R^d \to \mathcal{Y}\subset \R$. 
	Here we only consider \emph{symmetric} algorithms, \emph{i.e.}, $\A$ does not depend on the order of the sample points $Z_1,\dots,Z_n$. We also assume that for any $1\leq i \leq n$,
%
	
		\subparagraph{Assumptions.}
		
		To ease the reading of what follows, we provide simplified results under the following (somewhat restrictive) assumptions:
		\begin{itemize}
			\item \emph{Boundedness of $X$:}~\\
			Let us assume that there exists $0<B_X<+\infty$ such that
			\begin{align}\label{eq.norm}
			\forall\ 1\leq i\leq n,\quad \abs{X_i}_2 \leq B_X	,\ a.s. 	\tag{{\bf XBd}}
			\end{align}

			\item \emph{Boundedness of $Y$:}~\\
			Let us assume that there exists $0<B_{Y}<+\infty$ such that
			\begin{align}\label{assum.Y.boundedness}
			\forall\ 1\leq i\leq n,\quad \abs{Y_i} \leq B_{Y}	,\ a.s.	\tag{{\bf YBd}}
			\end{align}
			
			\item  \emph{Sub-Gaussianity of $Y$:}~\\
			Let us assume there exists $v>0$ such that
			\begin{align}\label{eq.subGaussian}
            \forall\ 1\leq i\leq n,\quad  \norm{Y_i - \E\croch{Y_i} }_q \leq 2e \sqrt{v} \sqrt{q}. \hfill \tag{\bf{SubG}}
            \end{align}
		\end{itemize}
		However	let us emphasize that most of the forthcoming results can be extended at the price of additional technicalities to the unbounded case (at least for instance to the Gaussian setting for $X$).

	
	The performance of algorithm $\A$ trained from $\D$ and evaluated at point $X$ is $c(\A(\D,X),Y)$, where $c(\cdot,\cdot):\ \Y\times \Y \to \R_+$ is a \emph{cost function}.
	The \emph{prediction error} of the estimator $\A(D)$ is the random variable depending on $\D$ given by
	\begin{equation}\label{eq.prediction.error}
	    \mathcal{L}_{P}\paren{\A(\D)} = \E_{(X,Y)\sim P}\croch{ c\paren{\A(\D,X), Y} }.
	\end{equation}
    
    In the sequel we let $\abs{\cdot}$ denote the absolute value in $\R$, $\eucl{\cdot}$ the Euclidean norm in $\R^d$, $\norm{\cdot}_{op}$ the operator norm over the set of $d\times d$ matrices $\mathcal{M}_d(\R)$, and $\norm{\cdot}_q$ the $L^q(\mathbb{P})$-norm for any $q\geq 1$ where $\P$ is a reference probability, \emph{i.e.}, $\norm{U}_q = \paren{\int \abs{U}^q \, d\P}^{1/q}$ for any real-valued random variable $U$.

	\subsection{A new notion of stability: $L^q$ stability}

    %
    %
    Our purpose is to bridge the gap between the weak notion of $L^1$ stability \citep[][Definition~3]{BE2002}, which only provides PAC polynomial generalisation bounds \citep[][Section 4.1]{BE2002}, and the stronger notion of uniform stability \citep[][Definition~6]{BE2002}, which leads to PAC exponential bounds yet may appear too restrictive \citep[][Section~3.1]{KN2002}.

    To this end the following definition generalises the $L^1$ stability to higher order moments.
	\begin{defn}[$L^q$ stability]\label{Def.Stab}
		Let  $\A$ denote any symmetric learning algorithm, and $c(\cdot,\cdot)$ be any cost function.
		Then for every $q\geq 1$, $\A$ is said $\gamma_q$-$L^q$ stable if there exists $\gamma_q>0$ such that  
		    \begin{align*}
    \forall \ 1\leq j\leq n,\quad	\mathcal{S}_q(\A,n)^q = 	\E\croch{ \abs{ c\paren{\A(\D,X),Y } - c\paren{\A( \tau_j(\D) , X),Y }  }^q } \leq \gamma_q^q ,
		\end{align*} 
		where the expectation is computed over $\D$ and $(X,Y)\sim P$, with $(X,Y)$ independent of $\D$, and $\tau_j(\D) = (Z_1,\dots,Z_{j-1},Z_{j+1},\dots,Z_n)$ is the sample $\D$ where $Z_j=(X_j,Y_j)$ has been removed.
	\end{defn}
	The above \autoref{Def.Stab}
	requires to bound the variation of $\A$ induced by removing one training point.
	This is in accordance with earlier definitions [\citealp{DeWa79}, \citealp{BE2002} and \citealp{EvgeniouPontilElisseeff_2004}].
	However, controlling high order moments provides more information on the distribution of 
	$c\paren{\A(\D,X),Y }$ than simply considering hypothesis stability, that is $L^q$ stability with $q=1$.
	Let us also mention that other notions of stability have been introduced, which replace one training point by an independent copy [\citealp{KN2002}, \citealp{kale2011cross} and \citealp{kumar2013near}].
Finally let us emphasise that uniform stability obviously implies $L^q$ stability for every $q\geq 1$.

	\subsection{Instances of stable learning algorithms}
	\label{sec.examples.stable.learning.algorithms}
	
    We now illustrate how the $L^q$ stability notion translates onto two learning algorithms: the $k$-nearest neighbors and the Ridge regression algorithms \citep[][Section 13.3 and Section 3.4]{friedman2001elements}.
	\paragraph{The $k$-nearest neighbors algorithm.}
For $1\leq k\leq n-1$, let $V_k(x)$ be the set of indices of the $k$ nearest neighbors ($k$NN) of $x$ among $X_1,\ldots,X_n$.
For binary classification, the $k$NN classifier is
	\begin{equation}\label{eq.knn.classifier}
	\A_{k}(\D;x)  =
	\begin{cases}
		 1 & \mbox{if } \sum_{j=1}^n Y_j \ind_{\acc{ j\in V_k(x)}} \geq k/2,\\
	 0 & \mbox{otherwise}.
	\end{cases}
	\end{equation}

\begin{prop}[\cite{DeWa79}, Eq.~(14)]\label{lem.kNN.stability}
With the above notation, for every $1\leq k\leq n-1$, $\A_k$ is $\gamma_1$-$L^1$ stable for the $0-1$ cost function $c(y,y^\prime) = \ind_{\acc{y\neq y^\prime}}$ with
\begin{align*}
\gamma_1 = \frac{4}{\sqrt{2\pi}}\frac{\sqrt{k}}{n}\enspace \cdot
\end{align*}
\end{prop}
	
	\begin{proof}[Proof of \autoref{lem.kNN.stability}]
	For every $1\leq j\leq n$, and using \autoref{Lemma : HOresult} for the last inequality,
		\begin{align*}
		\E\croch{ \abs{ c\paren{\A(\D)(X),Y } - c\paren{\A( \tau_j(\D) )(X),Y }  } } & =  \E\croch{ \abs{ \ind_{\acc{ \A_k(\D)(X)\neq Y}}  -  \ind_{\acc{ \A_k( \tau_j(\D) )(X)\neq Y}}  } } \\
		%
		%
		& =  \P\croch{   \A_k(\D)(X)\neq\A_k( \tau_j(\D) )(X) } \leq  \frac{4}{\sqrt{2\pi}}\frac{\sqrt{k}}{n} \enspace \cdot
		\end{align*}
	\end{proof}

	\paragraph{The Ridge regression algorithm.}
	
Let us recall that for any $\lambda>0$, the Ridge estimator is given by
\begin{align}\label{eq.def.Ridge.estimator}
\A_{\lambda}\paren{\D}= \underset{\beta\in \R^d}{\arg\min} \acc{ \frac{1}{n}\sum_{i=1}^n \paren{Y_i - \langle X_i,\beta\rangle_{\R^d}}^2 + \lambda \abs{\beta}_2^2 }=\frac{1}{n} \paren{ \widehat \Sigma + \lambda I_d }^{-1} X^T Y,
\end{align}
%
	where $\widehat \Sigma = X^TX/n = 1/n \sum_{i=1}^n X_i X_i^T$ denotes the empirical covariance matrix.
	Then,
		\begin{thm}\label{thm.Ridge.regression.stability}
		For any sample size $n>1$, $\eta\in (0,1)$, and $\lambda >\croch{\eta(n-1)}^{-1}$, let $\A_{\lambda}$ be given by Eq.~\eqref{eq.def.Ridge.estimator} and set $c(y,y^\prime) = \paren{y-y^\prime}^2$. 
		Then, assuming \eqref{eq.norm}, $\A_{\lambda}$ is $\gamma_q$-$L^q$ stable for any $q\geq 1$ with 
		\begin{align*}
		\gamma_q  & = 2 \norm{ Y}_{2q}^2 \frac{ B_X^2}{n\lambda} \paren{ 1  +  \frac{B_X^2+\lambda}{ \lambda (1 - \eta) }  }    \paren{ 1+ \frac{ B_X^2 }{ \lambda }}	 ,
		\end{align*}
		where $\gamma_q=+\infty$, if $ \norm{ Y  }_{2q} = +\infty$.
	\end{thm}
	The $L^q$ stability holds with the Ridge algorithm under the very mild assumption that $Y$ admits finite moments of order $q$ for some $q\geq 1$.
	Unsurprisingly, the stronger assumption of \citet[][Example~3]{BE2002} leads to a similar upper bound in terms of $L^q$ stability.

	The proof of \autoref{thm.Ridge.regression.stability} relies on the two following technical lemmas.

	\begin{lem}\label{lem.upper.bound.difference.estimated.vector}
		With the above notation, let us define $\eta \in (0,1)$ and $n$ satisfy $n\eta >1$.
		If \eqref{eq.norm} holds true, then for every $\lambda> B_X^2\paren{n\eta - 1}^{-1}$, it results
		\begin{align*}
		\eucl{ \A_{\lambda}\paren{\D}  - \A_{\lambda}\paren{\tau_j(\D)}  }  \leq \frac{ B_X}{n\lambda} \paren{ \abs{Y_j}  +  \frac{B_X^2+\lambda}{ \lambda (1 - \eta) } \croch{ \frac{1}{n-1} \sum_{i \neq j} \abs{Y_i } } } .
		\end{align*}
	\end{lem}
	
	\begin{proof}[Proof of  Lemma~\ref{lem.upper.bound.difference.estimated.vector}]
		Set $\widehat \Sigma^{(j)} = 1/(n-1)\sum_{i\neq j} X_iX_i^T$.
		\begin{align*}
		\eucl{ \A_{\lambda}\paren{\D}  - \A_{\lambda}\paren{\tau_j(\D)}  } 
		%
		%
		& \leq \eucl{ \frac{1}{n} \paren{ \widehat \Sigma + \lambda I_d }^{-1} \croch{ X^T Y-  (X^{(j)})^T Y^{(j)} } } \\
		& + \eucl{ \croch{ \frac{1}{n} \paren{ \widehat \Sigma + \lambda I_d }^{-1} - \frac{1}{n-1} \paren{ \widehat \Sigma^{(j)} + \lambda I_d }^{-1} } (X^{(j)})^T Y^{(j)} } \\
		& = T_1 + T_2 .
		\end{align*}	
		First, Lemma~\ref{lem.upper.bound.sym.positive.matrix} provides
		\begin{align*} 
		T_1 & = \eucl{ \frac{1}{n} \paren{ \widehat \Sigma + \lambda I_d }^{-1} \croch{ X^T Y-  (X^{(j)})^T Y^{(j)} } }  \leq \frac{1}{n} \norm{ \paren{ \widehat \Sigma + \lambda I_d }^{-1}}_{op} \eucl{   X_j Y_j } \leq   \frac{1}{n \lambda} \abs{ Y_j } \eucl{X_j}.
		\end{align*}
		Then, \eqref{eq.norm} yields
		\begin{align}
		T_1 &  \leq   \frac{1}{n \lambda} \abs{ Y_j } B_X . 
		\label{eq.T1}
		\end{align}
		Second, it is straightforward to observe that
		\begin{align*}
		T_2 
		%
		%
		& \leq  \frac{1}{n} \norm{  \paren{ \widehat \Sigma + \lambda I_d }^{-1} - \paren{ \frac{n-1}{n} \widehat \Sigma^{(j)} + \lambda I_d  - \frac{1}{n} \lambda I_d  }^{-1}  }_{op} \eucl{ \sum_{i \neq j}X_i Y_i } .
		\end{align*}
		Further, writing $\frac{n-1}{n}\widehat \Sigma^{(j)} + \lambda I_d = \widehat \Sigma + \lambda I_d + \frac{n-1}{n}\widehat \Sigma^{(j)} - \widehat \Sigma  $, we obtain
		\begin{align*}
		\paren{ \widehat \Sigma + \lambda I_d }^{-1} - \paren{ \frac{n-1}{n} \widehat \Sigma^{(j)} + \lambda I_d  - \frac{1}{n} \lambda I_d  }^{-1} = \paren{ \widehat \Sigma + \lambda I_d }^{-1} - \paren{\widehat \Sigma + \lambda I_d + B_j }^{-1} ,
		\end{align*}
		with $B_j =  \frac{n-1}{n}\widehat \Sigma^{(j)} - \widehat \Sigma - \frac{1}{n} \lambda I_d = -\paren{ X_jX_j^T+ \lambda I_d}/n$. Then, Lemma~\ref{lem.harville} and Lemma~\ref{lem.upper.bound.sym.positive.matrix} yield
		\begin{align*}
		& \norm{  \paren{ \widehat \Sigma + \lambda I_d }^{-1} - \paren{ \frac{n-1}{n} \widehat \Sigma^{(j)} + \lambda I_d  - \frac{1}{n} \lambda I_d  }^{-1}  }_{op}\\
		& \leq  \norm{  \paren{ \widehat \Sigma + \lambda I_d }^{-1} }_{op}^2 \norm{\paren{ X_jX_j^T+ \lambda I_d}/n}_{op} \norm{ \paren{ I_d + B_j \paren{ \widehat \Sigma + \lambda I_d }^{-1} }^{-1} }_{op} \\
		& \leq  \frac{ \eucl{X_j}^2+\lambda}{n\lambda^2} \norm{ \paren{ I_d + B_j \paren{ \widehat \Sigma + \lambda I_d }^{-1} }^{-1} }_{op} .
		\end{align*}
		Further assuming that for every $\eta \in (0,1)$, $n>\eta^{-1}$ and $	\lambda > \frac{B_X^2}{n\eta - 1}$, then Lemma~\ref{prop.bhatia} and \eqref{eq.norm} lead to
		\begin{align*}
		\norm{ \paren{ I_d + B_j \paren{ \widehat \Sigma + \lambda I_d }^{-1} }^{-1} }_{op} \leq \paren{1 - \frac{\norm{B_j}_{op}}{\lambda} }^{-1} \leq \paren{ 1 - \frac{B_X^2 + \lambda }{n\lambda} }^{-1} \leq \paren{ 1 - \eta }^{-1}  .
		\end{align*}
		%
		Using  \eqref{eq.norm}, this allows to conclude that
		\begin{align} \label{eq.T2}
		T_2 & \leq \frac{B_X^2+\lambda}{n \lambda^2} \frac{B_X}{ 1 - \eta } \croch{ \frac{1}{n-1} \sum_{i \neq j} \abs{Y_i } } .
		\end{align}
		Finally  gathering Eq.~\eqref{eq.T1} and~\eqref{eq.T2}, one obtains
		\begin{align*}
		\eucl{ \A_{\lambda}\paren{\D}  - \A_{\lambda}\paren{\tau_j(\D)}  }  \leq \frac{ B_X}{n\lambda} \paren{ \abs{Y_j}  +  \frac{B_X^2+\lambda}{ \lambda (1 - \eta) } \croch{ \frac{1}{n-1} \sum_{i \neq j} \abs{Y_i } } } .
		\end{align*}
	\end{proof}

\begin{lem}\label{lem.upper.bound.additive.term.quadratic.loss}
		With the above notation, let us define $\eta \in (0,1)$ and $n$ satisfy $n\eta >1$.
		If \eqref{eq.norm} holds true, then for every $\lambda> B_X^2\paren{n\eta - 1}^{-1}$, one has
\begin{align*}
    \abs{ 2Y - \A_{\lambda}\paren{\D;X}-  \A_{\lambda}\paren{\tau_j(\D);X} }  
	& \leq 2 \abs{Y} + B_X\paren{ \frac{1}{n\lambda} \sum_{i=1}^n \abs{Y_i }  + \frac{1}{(n-1)\lambda} \sum_{i\neq j} \abs{Y_i} } .
\end{align*}
\end{lem}

\begin{proof}[Proof of \autoref{lem.upper.bound.additive.term.quadratic.loss}]
	Combining the Cauchy-Schwarz inequality with \eqref{eq.norm} yields
	\begin{align*}
	 \abs{ 2Y - \A_{\lambda}\paren{\D;X}-  \A_{\lambda}\paren{\tau_j(\D);X} }  &\leq 2 \abs{Y} +  \abs{ <X,\A_{\lambda}\paren{\D}>_{\R^d}  }  + \abs{ <X,\A_{\lambda}\paren{\tau_j(\D) }>_{\R^d} }  \nonumber \\
    & \leq 2 \abs{Y} + B_X \paren{ \eucl{\A_{\lambda}\paren{\D}}  + \eucl{\A_{\lambda}\paren{\tau_j(\D)} } } \nonumber \\
	& \leq 2 \abs{Y} + B_X\paren{ \frac{1}{n\lambda} \sum_{i=1}^n \abs{Y_i }  + \frac{1}{(n-1)\lambda} \sum_{i\neq j} \abs{Y_i} } ,
	\end{align*}
	since Eq.~\eqref{eq.def.Ridge.estimator} and~\eqref{eq.norm} imply $ \eucl{	\A_{\lambda}\paren{\D} } \leq \frac{1}{n\lambda} \eucl{ \sum_{i=1}^n X_i Y_i} \leq B_X/(n\lambda)  \sum_{i=1}^n \abs{ Y_i}  $.

\end{proof}

	\begin{proof}[Proof of \autoref{thm.Ridge.regression.stability}]
		With $c(t(x),y) = \paren{ t(x)-y }^2$, any $q\geq 1$, the Cauchy-Schwarz inequality provides
		\begin{align*}
		\S_q\paren{\A_{\lambda}(\D)}	& =  \norm{	c\paren{ \A_{\lambda}\paren{\D;X}, Y } - c\paren{ \A_{\lambda}\paren{\tau_j(\D);X}, Y } }_{q} \\
		& \leq \norm{ \A_{\lambda}\paren{\D;X} - \A_{\lambda}\paren{\tau_j(\D);X} }_{2q}  \norm{	 2Y - \A_{\lambda}\paren{\D;X}-  \A_{\lambda}\paren{\tau_j(\D);X}  }_{2q} ,
		\end{align*}
		since $c(a,y) - c(b,y) = \paren{a-b} \paren{a+b-2y}$.
		
		Another use of the Cauchy-Schwarz inequality combined with $\A_{\lambda}\paren{\D;X} = \scal{\A_{\lambda}\paren{\D},X}$ and the independence between $(X,Y)$ and $\D$ leads to
		\begin{align}\label{ineq.upper.bound.stability.ridge.proof}
		\norm{ \A_{\lambda}\paren{\D;X} - \A_{\lambda}\paren{\tau_j(\D);X} }_{2q} & \leq \norm{\abs{X}_2}_{2q} \times 		\norm{\eucl{\A_{\lambda}\paren{\D} - \A_{\lambda}\paren{\tau_j(\D)} } }_{2q} \notag \\
		& \leq B_X \times 		\norm{\eucl{\A_{\lambda}\paren{\D} - \A_{\lambda}\paren{\tau_j(\D)} } }_{2q}  . 
		\end{align}
		Let us notice that  \autoref{lem.upper.bound.difference.estimated.vector} implies
		\begin{align}
		\norm{\eucl{\A_{\lambda}\paren{\D} - \A_{\lambda}\paren{\tau_j(\D)} } }_{2q}   \nonumber 
		%
		%
		& \leq \frac{ B_X}{n\lambda} \paren{ \norm{ Y_j }_{2q}  +  \frac{B_X^2+\lambda}{ \lambda (1 - \eta) } \norm{ \frac{1}{n-1} \sum_{i \neq j} \abs{Y_i } }_{2q} }  \nonumber\\
		& \leq \frac{ B_X}{n\lambda} \paren{ 1  +  \frac{B_X^2+\lambda}{ \lambda (1 - \eta) }  } \norm{ Y }_{2q} , \label{eq.Ridge.stability.terme1}
		\end{align}
		where $Y$ denotes an independent copy of the $Y_i$s.
		
		Likewise, \autoref{lem.upper.bound.additive.term.quadratic.loss} results in
		\begin{align*}
		& \norm{	 2Y - \A_{\lambda}\paren{\D;X}-  \A_{\lambda}\paren{\tau_j(\D);X}  }_{2q}  \\
		%
		%
		%
		& \leq 
		2 \norm{ Y}_{2q} + B_X^2 \paren{ \norm{ \frac{1}{n\lambda} \sum_{i=1}^n \abs{Y_i}  }_{2q} +  \norm{  \frac{1}{(n-1)\lambda} \sum_{i\neq j} \abs{Y_i }  }_{2q}  }  .
		\end{align*}
		Since the $Y_i$s are identically distributed, the triangular inequality gives
		\begin{align}\label{eq.Ridge.stability.terme2}
		& \norm{	 2Y - \A_{\lambda}\paren{\D;X}-  \A_{\lambda}\paren{\tau_j(\D);X}  }_{2q}   \leq 
		2 \norm{ Y}_{2q} \paren{ 1+ \frac{ B_X^2 }{\lambda} }  .
		\end{align}
		
		By combining Eq.~\eqref{ineq.upper.bound.stability.ridge.proof},~\eqref{eq.Ridge.stability.terme1} and~\eqref{eq.norm}, we obtain
		\begin{align*}
		& \norm{	c\paren{ \A_{\lambda}\paren{\D;X}, Y } - c\paren{ \A_{\lambda}\paren{\tau_j(\D);X}, Y } }_{q} \\
		& \leq \frac{ B_X^2}{n\lambda} \paren{ 1  +  \frac{B_X^2+\lambda}{ \lambda (1 - \eta) }  } \norm{ Y }_{2q} \times  2 \norm{ Y}_{2q} \paren{ 1+ \frac{ B_X^2 }{ \lambda }} \\
		& \leq 2 \norm{ Y}_{2q}^2 \frac{ B_X^2}{n\lambda} \paren{ 1  +  \frac{B_X^2+\lambda}{ \lambda (1 - \eta) }  }    \paren{ 1+ \frac{ B_X^2 }{ \lambda }} \cdot	
		\end{align*}
		
	\end{proof}

	\section{Deriving moment generalisation bounds from $L^q$ stability}
	\label{sec:moments}

Let us recall that the goal is to upper bound with high probability the discrepancy between the LoO estimator and the prediction error $\widehat{R}_1(\A,\D) - \L_P\paren{\A(\D)}$.
We now derive generalisation bounds in terms of moments for the LoO estimator of some learning algorithm $\A$. 
Recalling that the LoO estimator associated with $\A$ is
	\begin{equation*}
	\widehat{R}_1(\A,\D) = \frac{1}{n}\sum_{i=1}^n c(\A(\tau_i(\D),X_i),Y_i) ,
	%
	\end{equation*}
	where $\tau_i(\D) = (Z_1,\dots,Z_{i-1},Z_{i+1},\dots,Z_n)$, we now provide the main result of this section.
	It upper bounds the moments of $\widehat{R}_1(\A,\D) - \L_P\paren{\A(\D)}$.
The main steps of its proof follow \autoref{cor.moments.LoO.prediction.error} below.

\begin{thm}\label{thm.moments.LoO.prediction.error}
For any sample size $n\geq 2$ and any symmetric learning algorithm $\A$, let $\widehat{R}_1(\A,\D)$ and $\L_P\paren{\A(\D)}$ respectively denote the LoO estimator and the prediction error (see Eq.~\eqref{eq.prediction.error}).
Then there exists a numerical constant $0<\kappa\leq 1.271$ such that, for any $q\geq 2$, 
\begin{align*}
& \norm{ \widehat{R}_1(\A,\D) - \L_P\paren{\A(\D)}  - \paren{\E\croch{ \L_P\paren{\A(\D)} } -  \E\croch{ \widehat{R}_1(A,\D) }} }_q  \\
& \leq  \sqrt{\kappa q n}  \croch{ \sqrt{2  }  \,\mathcal{S}_q(\A,n)  +   4  \, \mathcal{S}_q(\A,n-1) } + \frac{2 \sqrt{\kappa q} }{\sqrt{n}}  \norm{ 
	c(\A(\tau_j(\D),X_j),Y_j) -  c(\A(\tau_j(\D),X_j^\prime),Y_j^\prime)  }_{q}   ,
\end{align*}
where $\S_q(\A,n)$ is given by \autoref{Def.Stab}. 
\end{thm}
This result implies that $\Rh_1(\A,\D)$ is a consistent estimator of 
$\L_P\paren{\A(\D)}$ in $L^q(\P)$ provided that $\S_q(\A,n) = o(1/\sqrt{n})$ as $n\to +\infty$.
For instance this holds true for the Ridge estimator with any $q\geq 2$ (\autoref{thm.Ridge.regression.stability}).

The proof of \autoref{thm.moments.LoO.prediction.error} heavily relies on the following generalisation of the Efron-Stein inequality \citep[see][Theorem~1 for Efron-Stein inequality]{BE2002}. 
\begin{prop}[\cite{boucheron2013concentration}, \cite{CMH}]
\label{cor.generalized.Efron.Stein}
Let $X_1,\ldots,X_n$ denote $n$ independent random variables and $Z=f(X_1,\ldots,X_n)$, where $f:\ \R^n\rightarrow \R$ is any Borel function.
With $Z_j'=f(X_1,\ldots,X_j',\ldots,X_n)$, where $X'_1,\ldots,X'_n$ are independent copies of the $X_i$s, there exists a universal constant $\kappa\leq 1.271$ such that for any $q\geq 2$,
\begin{align*}
\norm{ Z-\E Z }_q & \leq \sqrt{2\kappa q} \sqrt{\norm{\sum_{j=1}^n \paren{Z-Z_j^{\prime}}^2}_{q/2}} . 
	\end{align*}
\end{prop}
\autoref{cor.generalized.Efron.Stein} is a concentration result of $Z$ around its expectation.
This suggests a strategy to prove \autoref{thm.moments.LoO.prediction.error} that is based on the triangular inequality and the successive control of $\Rh_1(\A,\D)$ and $\L_P\paren{\A(\D)}$ around their expectations. This is done in the following two lemmas.
	\begin{lem}\label{thm.LoO.moments.stability}
    With the above notation, for any $q\geq 2$,
	    \begin{align*}
	        & \norm{ \Rh_1(\A,\D) - \E\croch{\Rh_1(\A,\D)} }_q \\
	        & \leq 2 \sqrt{\kappa q} \croch{ 2  \sqrt{  n}  \, \mathcal{S}_q(\A,n-1) + \frac{1}{\sqrt{n}}  \norm{ 
	        	c(\A(\tau_j(\D),X_j),Y_j) -  c(\A(\tau_j(\D),X_j^\prime),Y_j^\prime)  }_{q}  } .
	    \end{align*}
    \end{lem}

\begin{proof}[Proof of \autoref{thm.LoO.moments.stability}]
First, note that \autoref{cor.generalized.Efron.Stein} gives
	\begin{align*}
	\norm{ \Rh_1(\A,\D) - \E\croch{\Rh_1(\A,\D)} }_q %
	& \leq \sqrt{2\kappa q} \sqrt{\norm{\sum_{j=1}^n \paren{ \Rh_1(\A,\D) - \Rh_1(\A,\D^j) }^2}_{q/2}} ,
	\end{align*}
	where $\D^j = (Z_1,\ldots,Z_j^\prime,\ldots,Z_n)$ and $Z_j^\prime$ is an independent copy of $Z_j$ for any $j$.
	Moreover,
	\begin{align*}
	& \norm{\sum_{j=1}^n \paren{ \Rh_1(\A,\D) - \Rh_1(\A,\D^j) }^2}_{q/2} \\
	%
	%
	%
	& \leq \sum_{j=1}^n  2 \paren{ \frac{1}{n} \sum_{i\neq j} \norm{ c(\A(\tau_i(\D),X_i),Y_i) - c(\A(\tau_i(\D^j),X_i),Y_i)   }_{q}}^2 \\
	& \quad + \frac{2}{n}  \norm{ c(\A(\tau_j(\D),X_j),Y_j) -  c(\A(\tau_j(\D),X_j^\prime),Y_j^\prime)     }_{q}^2 \\
	& =  2 n  \paren{ \frac{n-1}{n}}^2 4\, \mathcal{S}_q^2(\A,n-1) + \frac{2}{n}  \norm{ 
	c(\A(\tau_j(\D),X_j),Y_j) -  c(\A(\tau_j(\D),X_j^\prime),Y_j^\prime)  }_{q}^2 \\
		& \leq  8 n  \, \mathcal{S}_q^2(\A,n-1) + \frac{2}{n}  \norm{ 
			c(\A(\tau_j(\D),X_j),Y_j) -  c(\A(\tau_j(\D),X_j^\prime),Y_j^\prime)  }_{q}^2  ,
		\end{align*}
		and the proof ends by taking the square root on both sides of the inequality and using $\sqrt{a+b} \leq \sqrt{a} + \sqrt{b}$, for every $a,b\geq0$.
\end{proof}

\begin{lem}\label{thm.prediction.error.stability}
Let $\L_P\paren{ \A(\D) }$ denote the prediction error given by Eq.~\eqref{eq.prediction.error}.
Then for any $q\geq 2$, 
\begin{align*}
    \norm{ \L_P\paren{\A(\D)} - \E\croch{ \L_P\paren{\A(\D)} } }_q 
    %
    & \leq \sqrt{2 \kappa q} \sqrt{ n } \,\mathcal{S}_q(\A,n)  .
\end{align*}
\end{lem}
\begin{proof}
	The proof is similar to that of \autoref{thm.LoO.moments.stability}.
	\begin{align*}
	\norm{ \L_P\paren{\A(\D)} - \E\croch{ \L_P\paren{\A(\D)} } }_q  
	& \leq \sqrt{2 \kappa q} \sqrt{ \norm{ \sum_{j=1}^n \croch{ \L_P\paren{\A(\D)} - \L_P\paren{\A(\tau_j(\D) ) } }^2 }_{q/2} } \\
	& \leq \sqrt{2 \kappa q} \sqrt{ n \norm{  \croch{ \L_P\paren{\A(\D)} - \L_P\paren{\A(\tau_j(\D) ) } }^2 }_{q/2} } \\
	& \leq \sqrt{2 \kappa q} \sqrt{ n \norm{  \croch{ c\paren{\A(\D,X),Y} - c\paren{\A(\tau_j(\D),X ) , Y } }^2 }_{q/2} } \\
	& = \sqrt{2 \kappa q} \sqrt{ n } \norm{  \croch{ c\paren{\A(\D,X),Y} - c\paren{\A(\tau_j(\D),X ) , Y } }^2 }_{q}  \\
	& = \sqrt{2 \kappa q} \sqrt{ n } \,\mathcal{S}_q(\A,n) .
	\end{align*}
\end{proof}	

\begin{proof}[Proof of \autoref{thm.moments.LoO.prediction.error}]
The triangular inequality and Lemmas~\ref{thm.LoO.moments.stability} and~\ref{thm.prediction.error.stability} lead to
\begin{align*}
&    \norm{ \L_P\paren{\A(\D)} - \widehat{R}_1(A,\D)  - \paren{\E\croch{ \L_P\paren{\A(\D)} } -  \E\croch{ \widehat{R}_1(A,\D) }}  }_q \\
& \leq  \norm{ \L_P\paren{\A(\D)} - \E\croch{\L_P\paren{\A(\D)}} }_q 
 + \norm{ \E\croch{ \widehat{R}_1(A,\D) } - \widehat{R}_1(A,\D) }_q \\
& \leq  \sqrt{\kappa q n}  \croch{ \sqrt{2  }  \,\mathcal{S}_q(\A,n)  +   4  \, \mathcal{S}_q(\A,n-1) }+ \frac{2 \sqrt{\kappa q} }{\sqrt{n}}  \norm{ 
		c(\A(\tau_j(\D),X_j),Y_j) -  c(\A(\tau_j(\D),X_j^\prime),Y_j^\prime)  }_{q}   .
\end{align*}
\end{proof}

\subsection{Application to Ridge regression}
We now extend the result of \autoref{thm.moments.LoO.prediction.error} to the case of the Ridge regression algorithm.
\begin{cor}[Ridge: bounding the moments]
	\label{cor.moments.LoO.prediction.error}
	With the notation of \autoref{thm.moments.LoO.prediction.error}, for any sample size $n>2$, $\eta\in (0,1)$, and $\lambda >\croch{\eta(n-2)}^{-1}$, let $\A_{\lambda}(\D)$ denote the Ridge estimator from Eq.~\eqref{eq.def.Ridge.estimator}.
	Then, assuming \eqref{eq.norm}, for any $q\geq 2$, 
	\begin{enumerate}
		\item[(i)]
	\begin{align}
	& \norm{ \widehat{R}_1(\A_{\lambda},\D) - \L_P\paren{\A_{\lambda}(\D)} - \E\croch{ \widehat{R}_1(\A_{\lambda},\D) - \L_P\paren{\A_{\lambda}(\D)} } }_q \notag \\
	& \leq  \frac{\sqrt{q}}{\sqrt{n}}   \paren{  \Gamma_{\lambda,1} \norm{Y}_q^2  + \Gamma_{\lambda,2} \norm{Y}_{2q}^2 } ,\label{eq.coro.centered}
	\end{align}

		\item[(ii)] 
		\begin{align} \label{eq.coro.uncentered}
		&	\norm{ \widehat{R}_1(\A_{\lambda},\D) - \L_P\paren{\A_{\lambda}(\D)} }_q  \leq \frac{\sqrt{q}}{\sqrt{n}}   \paren{  \Gamma_{\lambda,1} \norm{Y}_q^2  + \Gamma_{\lambda,2} \norm{Y}_{2q}^2 }  +  \frac{ \Gamma_{\lambda,3}}{n}	\norm{ Y}_{2q}^2 ,
		\end{align}	
\end{enumerate}
		where $\Gamma_{\lambda,1} = 8 \sqrt{\kappa}B_X^2/\lambda$, $\Gamma_{\lambda,2} = 2\sqrt{\kappa}B_X^2/\lambda \croch{ (8+ \sqrt{2})  \paren{ 1  +  \frac{B_X^2+\lambda}{ \lambda (1 - \eta) }  }    \paren{ 1+ \frac{ B_X^2 }{ \lambda }}	 +  \frac{4 B_X^2}{ \lambda } }$, and $\Gamma_{\lambda,3} = \frac{  2 B_X^2}{\lambda} \paren{ 1  +  \frac{B_X^2+\lambda}{ \lambda (1 - \eta) }  }    \paren{ 1+ \frac{ B_X^2 }{ \lambda }}$.
\end{cor}
The first Eq.~\eqref{eq.coro.centered} is a direct consequence of
\autoref{thm.moments.LoO.prediction.error}. 
We recall that our goal is to quantify the discrepancy between the LoO estimator and the prediction error $\widehat{R}_1(\A_{\lambda},\D) - \L_P\paren{\A_{\lambda}(\D)} $. This justifies introducing Eq.~\eqref{eq.coro.uncentered}.

With the Ridge estimator, the convergence rate of $\Rh_1(\A_{\lambda},\D)$ towards $\L_P\paren{\A_{\lambda}(\D)}$ is of order $1/\sqrt{n}$ for any $q\geq 2$.
The upper bound highlights that the distribution of $Y$ influences the convergence rate, which deteriorates as $\norm{Y}_q$ and $\norm{Y}_{2q}$ grow. In particular, the dependence of the rate with respect to $q$ is strictly $ \sqrt{q}$ whenever $Y$ is almost surely bounded as in \eqref{assum.Y.boundedness}. Under the weaker assumption that $Y$ is sub-Gaussian, an additional terms will emerge 
Note also that this bound allows to derive PAC polynomial bounds by use of Markov-type inequalities.

The proof of \autoref{cor.moments.LoO.prediction.error} relies on the following proposition.
\begin{prop}\label{prop.upper.bound.variance.term}
	Let $\A_{\lambda}(\D)$ denote the Ridge estimator from Eq.~\eqref{eq.def.Ridge.estimator} for every $\lambda>0$, and assume \eqref{eq.norm} holds true. Then, we have
\begin{align*}
\norm{ 
	c(\A_{\lambda}(\tau_j(\D),X_j),Y_j) -  c(\A_{\lambda}(\tau_j(\D),X_j^\prime),Y_j^\prime)  }_{q}  & \leq \frac{4 B_X^2}{ \lambda } \norm{Y}_q^2 +  \frac{4 B_X^4}{ \lambda^2 } \norm{Y}_{2q}^2  .
\end{align*}
\end{prop} 

\begin{proof}[Proof of \autoref{prop.upper.bound.variance.term}]
	From \eqref{eq.norm}, it comes
	\begin{align*}
	\abs{ \scal{ \A_{\lambda}(\tau_j(\D), X_j-X_j^\prime} } \leq 2 B_X \eucl{\A_{\lambda}(\tau_j(\D) } ,
	\end{align*} 
	and
	\begin{align*}
	\abs{ Y_j - \scal{ \A_{\lambda}(\tau_j(\D), X_j } + Y_j^\prime - \scal{ \A_{\lambda}(\tau_j(\D), X_j^\prime} } \leq \abs{Y_j} + \abs{Y_j^\prime} + 2B\eucl{\A_{\lambda}(\tau_j(\D)} .
	\end{align*}
	Since Eq.~\eqref{eq.def.Ridge.estimator} and~\eqref{eq.norm} imply $\eucl{\A_{\lambda}(\tau_j(\D)}\leq  \frac{B_X}{(n-1)\lambda } \sum_{i\neq j} \abs{Y_i } $, the independence between $Y_j$, $Y_j^\prime$ and $\acc{Y_i}_{i\neq j}$ provides
	\begin{align*}
	& \norm{ c\paren{\A_{\lambda}(\tau_j(\D),X_j),Y_j} - c\paren{\A_{\lambda}(\tau_j(\D),X_j^\prime),Y_j^\prime} }_q \\
	& \leq 2 B_X^2 \norm{  \frac{1}{(n-1)\lambda} \sum_{i\neq j} \abs{Y_i} }_q \norm{ \abs{Y_j}+ \abs{Y_j^\prime}}_q + 4 B_X^4 \norm{ \paren{ \frac{1}{(n-1)\lambda} \sum_{i\neq j} \abs{Y_i} }^2 }_q \\
	& \leq \frac{4 B_X^2}{ \lambda } \norm{Y}_q^2 +  \frac{4 B_X^4}{ \lambda^2 } \norm{Y}_{2q}^2  ,
	\end{align*}
	where $Y$ denotes an independent copy of the $Y_i$s and $Y_j^\prime$.

\end{proof}

\begin{proof}[Proof of \autoref{cor.moments.LoO.prediction.error}]
	To prove claim $(i)$, note that \autoref{thm.moments.LoO.prediction.error} and \autoref{prop.upper.bound.variance.term} lead to
	\begin{align*}
	& \norm{ \widehat{R}_1(\A_{\lambda},\D) - \L_P\paren{\A_{\lambda}(\D)} - \E\croch{ \widehat{R}_1(\A_{\lambda},\D) - \L_P\paren{\A_{\lambda}(\D)} } }_q  \\
	   & \leq  \sqrt{2 \kappa q} \sqrt{ n }  \,\mathcal{S}_q(\A_{\lambda},n)   + 4  \sqrt{\kappa q}   \sqrt{  n}  \, \mathcal{S}_q(\A_{\lambda},n-1)\\
	&\quad +   \frac{2 \sqrt{\kappa q}}{\sqrt{n}}  \norm{ 
		c(\A_{\lambda}(\tau_j(\D),X_j),Y_j) -  c(\A_{\lambda}(\tau_j(\D),X_j^\prime),Y_j^\prime)  }_{q}    \\
	%
		& \leq  2 (8+ \sqrt{2})\sqrt{ \kappa q} \sqrt{ n }  \, \norm{ Y}_{2q}^2 \frac{ B_X^2}{n\lambda} \paren{ 1  +  \frac{B_X^2+\lambda}{ \lambda (1 - \eta) }  }    \paren{ 1+ \frac{ B_X^2 }{ \lambda }}	\\
		&\quad +   \frac{2 \sqrt{\kappa q}}{\sqrt{n}} \paren{ \frac{4 B_X^2}{ \lambda } \norm{Y}_q^2 +  \frac{4 B_X^4}{ \lambda^2 } \norm{Y}_{2q}^2 }  \\
		& =  \frac{2 \sqrt{\kappa q}}{\sqrt{n}} \frac{ B_X^2}{\lambda}   \croch{ (8+ \sqrt{2}) \norm{ Y}_{2q}^2 \paren{ 1  +  \frac{B_X^2+\lambda}{ \lambda (1 - \eta) }  }    \paren{ 1+ \frac{ B_X^2 }{ \lambda }}	 +   4 \norm{Y}_q^2 +  \frac{4 B_X^2}{ \lambda } \norm{Y}_{2q}^2 } \\
		& =  \frac{2 \sqrt{\kappa q}}{\sqrt{n}} \frac{ B_X^2}{\lambda}   \paren{   4 \norm{Y}_q^2  + \croch{ (8+ \sqrt{2})  \paren{ 1  +  \frac{B_X^2+\lambda}{ \lambda (1 - \eta) }  }    \paren{ 1+ \frac{ B_X^2 }{ \lambda }}	 +  \frac{4 B_X^2}{ \lambda } } \norm{Y}_{2q}^2 } .
	\end{align*}
	
	To prove claim $(ii)$, it only remains to notice that $ \abs{ \E\croch{ \L_P\paren{\A_{\lambda}(\D)} } -  \E\croch{ \widehat{R}_1(A,\D) }  } \leq \S_q(\A_{\lambda},n) $ by Jensen's inequality.
It results
\begin{align*}
&	\norm{ \widehat{R}_1(\A_{\lambda},\D) - \L_P\paren{\A_{\lambda}(\D)} }_q  \\
& \leq \frac{2 \sqrt{\kappa q}}{\sqrt{n}} \frac{ B_X^2}{\lambda}   \paren{   4 \norm{Y}_q^2  + \croch{ (8+ \sqrt{2})  \paren{ 1  +  \frac{B_X^2+\lambda}{ \lambda (1 - \eta) }  }    \paren{ 1+ \frac{ B_X^2 }{ \lambda }}	 +  \frac{4 B_X^2}{ \lambda } } \norm{Y}_{2q}^2 } \\
& \quad +  \frac{  2 B_X^2}{n\lambda} \paren{ 1  +  \frac{B_X^2+\lambda}{ \lambda (1 - \eta) }  }    \paren{ 1+ \frac{ B_X^2 }{ \lambda }}	\norm{ Y}_{2q}^2 .
\end{align*}	
	
\end{proof}


	\section{PAC exponential inequalities for the Leave-one-Out estimator}
	\label{sec:expo}
	
	We now state our final results for the LoO estimator. The key ingredient is the following proposition, which establishes a link between moment inequalities and PAC exponential inequalities.
	
\begin{prop}[\citet{CMH}, Proposition~D.1]
\label{prop.moment.exponential}
  Let $X$ denote a real valued random variable, and assume there exist $C>0$, $\lambda_1,\ldots,\lambda_N>0$, and $\alpha_1,\ldots,\alpha_N>0$ ($N\in\N^*$) such that for any $q\geq q_0$,
$  \E\croch{ \abs{X}^q } \leq C \paren{ \sum_{i=1}^N \lambda_i q^{\alpha_i} }^q $.
Then for every $x>0$,
\begin{equation*}
  \P\croch{ \abs{X} > \sum_{i=1}^N \lambda_i \paren{ \frac{e x }{\min_j \alpha_j} }^{\alpha_i} } \leq C e^{q_0 \min_j \alpha_j}  \cdot e^{-x} . \hfill
\end{equation*}
\end{prop}

	The following two final results are our most refined PAC exponential inequalities for the LoO estimator, and follow from \autoref{cor.moments.LoO.prediction.error}
(derived for the Ridge algorithm) combined with \autoref{prop.moment.exponential} where $q_0=2$, $C=1$ and $\min_j \alpha_j = 1/2$.
\begin{thm}\label{thm.exponential.inequality.LoO.prediction.error1}
With the setting of \autoref{cor.moments.LoO.prediction.error} and assuming \eqref{assum.Y.boundedness}, we have for every $x>0$, with probability at least $1- e \cdot e^{-x}$, 
\begin{equation}\label{eq.LoO.prediction.error.exponential.ineq.bounded}
    \abs{  \Rh_1(\A_{\lambda},\D) - \L_P\paren{\A_{\lambda}(\D)} } \leq \sqrt{ \frac{2ex}{n} } B_Y^2 \, \paren{\Gamma_{\lambda,1} + \Gamma_{\lambda,2} + \Gamma_{\lambda,3}} ,
\end{equation}
where  
\begin{align*}
    \Gamma_{\lambda,1} & = 8 \sqrt{\kappa}B_X^2/\lambda ,\\
    \Gamma_{\lambda,2} & = 2\sqrt{\kappa}B_X^2/\lambda \croch{ (8+ \sqrt{2})  \paren{ 1  +  \frac{B_X^2+\lambda}{ \lambda (1 - \eta) }  }    \paren{ 1+ \frac{ B_X^2 }{ \lambda }}	 +  \frac{4 B_X^2}{ \lambda } }, \\
    \Gamma_{\lambda,3} & = \frac{  2 B_X^2}{\lambda} \paren{ 1  +  \frac{B_X^2+\lambda}{ \lambda (1 - \eta) }  }    \paren{ 1+ \frac{ B_X^2 }{ \lambda }} .
\end{align*}
\end{thm}
This result establishes with high probability that the LoO estimator is $1/\sqrt{n}$ close to the prediction error when applied to the Ridge regression estimator.

This rate of convergence is preserved when \eqref{assum.Y.boundedness} is relaxed to \eqref{eq.subGaussian} as shown below.
\begin{thm}
\label{thm.exponential.inequality.LoO.prediction.error2}
With the setting of \autoref{cor.moments.LoO.prediction.error} and assuming \eqref{eq.subGaussian}, we have for every $x>0$, with probability at least $1- e \cdot e^{-x}$
\begin{equation}\label{eq.LoO.prediction.error.exponential.ineq.subGaussian}
    \abs{ \Rh_1(\A_{\lambda},\D) - \L_P\paren{\A_{\lambda}(\D)} } \leq \frac{ M_1 ( \E\croch{Y} )^2  \sqrt{x} + M_2 \, v \,x^{3/2}  }{ \sqrt{n} } \enspace ,
\end{equation}
where $M_1 = 2\sqrt{2e}\paren{\Gamma_{\lambda,1}+\Gamma_{\lambda,2}+\Gamma_{\lambda,3}}$, and $M_2 = 16 e^2 (2e)^{3/2}  \paren{\Gamma_{\lambda,1}+\Gamma_{\lambda,2}+\Gamma_{\lambda,3}} $.
\end{thm}
Both Eqs.~\eqref{eq.LoO.prediction.error.exponential.ineq.bounded} and~\eqref{eq.LoO.prediction.error.exponential.ineq.subGaussian} lead to deviations for the LoO estimator of order $1/\sqrt{n}$, which is similar to that of \citet[][Theorem 18, Eq.~18]{BE2002}, but derived under weaker assumptions (in particular $L^q$ stability instead of uniform stability).  
Note also that relaxing \eqref{assum.Y.boundedness} to \eqref{eq.subGaussian} results in an additional term of magnitude $x^{3/2}$.

	\begin{proof}[Proof of \autoref{thm.exponential.inequality.LoO.prediction.error1}]
Using Assumption~\eqref{assum.Y.boundedness} and \autoref{cor.moments.LoO.prediction.error}, apply \autoref{prop.moment.exponential} with $N=1$, $\alpha_1=1/2$ and $\lambda_1 = B_Y^2 \paren{\Gamma_{\lambda,1} + \Gamma_{\lambda,2} + \Gamma_{\lambda,3}}/\sqrt{n}$ to obtain Eq.~\eqref{eq.LoO.prediction.error.exponential.ineq.bounded}.
\end{proof}
	
\begin{proof}[Proof of \autoref{thm.exponential.inequality.LoO.prediction.error2}]
	From Assumption~\eqref{eq.subGaussian}, the triangular inequality yields
	\begin{align*}
	    \norm{Y}_q^2 & \leq 2 \paren{\E\croch{Y} }^2  + 2 \norm{Y - \E\croch{Y} }_q^2
 \leq 2 \paren{\E\croch{Y} }^2 + 2 (2e)^2 v q  \\
    \norm{Y}_{2q}^2 & \leq 2 \paren{\E\croch{Y} }^2 + 2 \norm{Y - \E\croch{Y} }_{2q}^2 
        \leq 2 \paren{\E\croch{Y} }^2 + 4 (2e)^2 v q .
	\end{align*}
	To simplify the derivation, we use $\norm{Y}_q^2 \leq  2 \paren{\E\croch{Y} }^2 + 4 (2e)^2 v q $ and $1/n \leq 1/\sqrt{n}$.
    From \autoref{cor.moments.LoO.prediction.error},
    \begin{align*}
        \norm{ \Rh_1(\A,\D) - \L_P\paren{\A(\D)}  }_q 
        & \leq \frac{\sqrt{q}}{\sqrt{n}} \paren{\Gamma_{\lambda,1} + \Gamma_{\lambda,2} + \Gamma_{\lambda,3}} \croch{  2 \paren{\E\croch{Y} }^2 + 4 (2e)^2 v q } \\
        & \leq \frac{\sqrt{q}}{\sqrt{n}} \paren{\Gamma_{\lambda,1} + \Gamma_{\lambda,2} + \Gamma_{\lambda,3}} \croch{ 2 \paren{\E\croch{Y} }^2 + 4 (2e)^2 v q  }
    \end{align*}
    To achieve the proof, it only remains to apply \autoref{prop.moment.exponential} with $N=2$, $(\alpha_1,\alpha_2)=(1/2,3/2)$, and
    \begin{align*}
        \lambda_1 & = 2 \sqrt{2e} \paren{\Gamma_{\lambda,1} + \Gamma_{\lambda,2} + \Gamma_{\lambda,3}} \paren{\E\croch{Y} }^2 ,\\
        \lambda_2 & = 16 e^2  (2e)^{3/2} \paren{\Gamma_{\lambda,1} + \Gamma_{\lambda,2} + \Gamma_{\lambda,3}} v.
    \end{align*}

	\end{proof}

	
	\section{Perspectives}	\label{sec:discu}
	
	We introduced a new stability notion---$L^q$ stability---which generalises existing hypothesis stability while remaining weaker than uniform stability.
	It provides PAC exponential generalisation bounds similar to those originally derived under the stronger uniform stability assumption.
	This has been achieved using a new generic strategy relying on moment inequalities.
	
	In the present paper, this strategy has been successfully applied to the Ridge regression algorithm.
	A natural next step is to explore the collection of learning algorithms falling into the scope of our $L^q$ stability notion.
	
	From both practical and theoretical perspectives, it is crucial to provide a thorough analysis of CV. On the basis of our generic analysis of LoO, we intend to investigate other CV procedures such as $V$-fold CV, Leave-$v$-Out, and so on.
	

	
	\appendix
	
	\section{Technical results}\label{sec:appendix}
	
	\begin{lem}[\cite{DeWa79}, Eq.~ (14)]
		\label{Lemma : HOresult}
		For any $1\leq k \leq n$, let $\A_k$ defined by Eq.~\eqref{eq.knn.classifier}, and let $Z_1,\ldots,Z_n$ denote $n$ independent and identically distributed random variables such that for any $1\leq i\leq n$, $Z_i = (X_i,Y_i) \sim P$. Then for any $1\leq j\leq n$, 
		\begin{equation*}
			\P\croch{ \A_k(\D;X) \neq \A_k(\tau_j(\D);X) } \leq \frac{4}{\sqrt{2\pi}}\frac{\sqrt{k}}{ n} \enspace \cdot
		\end{equation*}
	\end{lem}
	
	\begin{lem}[\cite{henderson1981deriving}, Eq.~(18)] \label{lem.harville}
		Let $A$ and $B$ denote two symmetric matrices in $\mathcal{M}_d(\R)$ for some integer $d>0$ such that $A$ and $A+B$ are invertible.
		Then,
		\begin{equation*}
		 A^{-1} - \paren{A+B}^{-1} = A^{-1} B A^{-1} \paren{I_d + B A^{-1} }^{-1} .
		\end{equation*}
		
	\end{lem}
	
	\begin{lem}\label{lem.upper.bound.sym.positive.matrix}
		Let $M\in \mathcal{M}_d(\R)$ be any symmetric positive semidefinite matrix.
		Then for every $\lambda>0$, 
		\begin{equation*}
		\norm{ \paren{M + \lambda I_d}^{-1} }_{op} \leq \frac{1}{\lambda} \enspace\cdot
		\end{equation*}
	\end{lem}

	\begin{lem}[\cite{bhatia2013matrix}, Theorem~VIII.3.1] \label{prop.bhatia}
Let $D \in\mathcal{M}_d(\R)$ be a diagonal matrix and $M \in\mathcal{M}_d(\R)$ denote any matrix.
Then,
\begin{equation*}
\max_{1\leq j \leq d} \min_{1\leq i\leq d} \abs{ \sigma_i(D) - \sigma_j(M)} \leq \norm{D-M}_{op} ,
\end{equation*}
where $\sigma_1(N)\geq \ldots\geq \sigma_d(N)$ denote the $d$ eigenvalues of matrix $N$ in nonincreasing order, and $\norm{\cdot}_{op}$ is the operator norm.
	\end{lem}


	
	\bibliographystyle{abbrvnat}
	
	\bibliography{biblio}
	
	
\end{document}